\documentclass[conference]{IEEEtran}
\IEEEoverridecommandlockouts

\usepackage[switch]{lineno}

\usepackage{cite}
\usepackage{amsmath,amssymb,amsfonts}
\usepackage{algorithmic}
\usepackage{graphicx}
\usepackage{textcomp}
\usepackage{xcolor}
\usepackage{float}
\def\BibTeX{{\rm B\kern-.05em{\sc i\kern-.025em b}\kern-.08em
    T\kern-.1667em\lower.7ex\hbox{E}\kern-.125emX}}
    
\usepackage{enumitem}
\setlist[itemize]{itemsep=2pt}
\setlist[enumerate]{itemsep=2pt}
\usepackage[utf8]{inputenc} 
\usepackage[T1]{fontenc}  
\usepackage{hyperref}    
\usepackage{url}      
\usepackage{booktabs}    
\usepackage{multicol}
\usepackage{nicefrac}    
\usepackage{microtype}   
\usepackage{wrapfig}
\usepackage{subcaption}
\usepackage{multirow}

\makeatletter

\begin{document}

\title{Exploring Post-Training Quantization of Protein Language Models}

\author{\IEEEauthorblockN{Shuang Peng\IEEEauthorrefmark{2}, Fei Yang\IEEEauthorrefmark{2}$^*$\thanks{$^*$Fei Yang is the corresponding author.}, Ning Sun\IEEEauthorrefmark{2}, Sheng Chen\IEEEauthorrefmark{2}, Yanfeng Jiang\IEEEauthorrefmark{3} and
Aimin Pan\IEEEauthorrefmark{2}}
\IEEEauthorblockA{\IEEEauthorrefmark{2}Zhejiang Lab, Hangzhou, China\\
\IEEEauthorrefmark{3}Nankai University, Tianjin, China\\
\{pengs, yangf, sunning, scucs, panaimin\}@zhejianglab.com,
yfjiang@mail.nankai.edu.cn}}

\maketitle

\begin{abstract}

Recent advancements in unsupervised protein language models (ProteinLMs), like ESM-1b~\cite{rives2019biological} and ESM-2~\cite{lin2022language}, have shown promise in different protein prediction tasks. However, these models face challenges due to their high computational demands, significant memory needs, and latency, restricting their usage on devices with limited resources. To tackle this, we explore post-training quantization (PTQ) for ProteinLMs, focusing on ESMFold~\cite{lin2022language}, a simplified version of AlphaFold~\cite{jumper2021highly} based on ESM-2 ProteinLM.
Our study is the first attempt to quantize all weights and activations of ProteinLMs. We observed that the typical uniform quantization method performs poorly on ESMFold, causing a significant drop in TM-Score when using 8-bit quantization. We conducted extensive quantization experiments, uncovering unique challenges associated with ESMFold, particularly highly asymmetric activation ranges before Layer Normalization, making representation difficult using low-bit fixed-point formats.
To address these challenges, we propose a new PTQ method for ProteinLMs, utilizing piecewise linear quantization for asymmetric activation values to ensure accurate approximation. We demonstrated the effectiveness of our method in protein structure prediction tasks, demonstrating that ESMFold can be accurately quantized to low-bit widths without compromising accuracy. Additionally, we applied our method to the contact prediction task, showcasing its versatility. In summary, our study introduces an innovative PTQ method for ProteinLMs, addressing specific quantization challenges and potentially leading to the development of more efficient ProteinLMs with significant implications for various protein-related applications.
\end{abstract}


\section{Introduction}
Protein is the fundamental building block of life, carrying out the majority of cellular activities. For a long time, scientists have been investigating the mechanisms underlying the structural and functional properties of proteins. The canonical \emph{sequence-structure-function} relationship has enabled the success of sequence-based machine learning methods, which can infer protein structure and function from amino acid sequences~\cite{hu2022exploring}. Recently, protein language models (ProteinLMs) have emerged as the state-of-the-art method for protein function and fitness predictions~\cite{rao2019evaluating,rives2021biological,lin2022language,fang2022helixfold_single,OmegaFold}. ProteinLMs are inspired by natural language models, where a string of amino acid residues is treated as continuous words in a sentence. After being pre-trained on hundreds of millions of amino acid sequences, ProteinLMs acquire the representation abilities to predict protein properties. This self-supervised pre-training approach has proven to be effective in a variety of protein-related tasks, including 3D structure and function prediction.

\begin{figure*}[t]
  \centering
  \includegraphics[width = 0.9\textwidth]{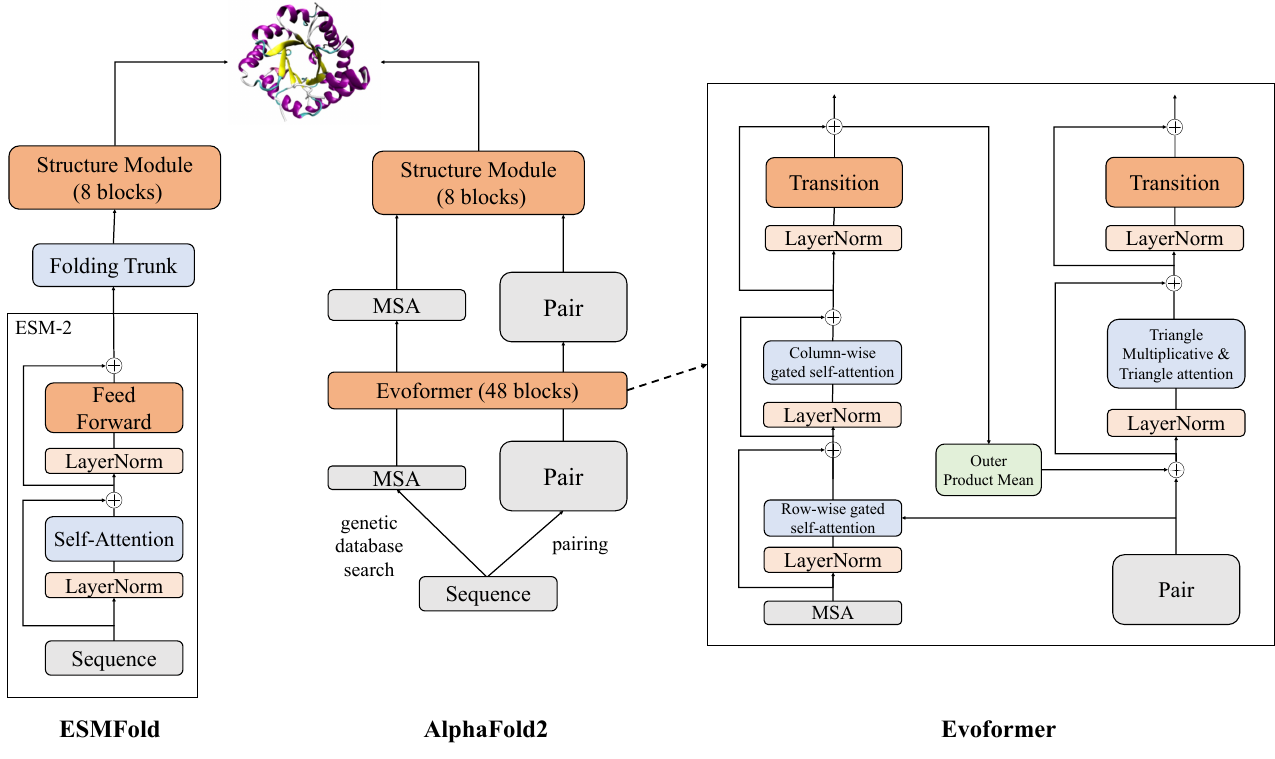}
  \caption{Core modules of ESM-2, ESMFold and Alphafold2. The Folding Trunk in ESMFold is a simplified single-sequence version of the EvoFormer described in AlphaFold2~\cite{lin2022language}.}
  \label{plms}
\end{figure*}

ESM-2~\cite{lin2022language} represents the cutting-edge ProteinLMs for various structure prediction tasks. ESMFold, a modified AlphaFold Evoformer and Structure Module~\cite{lin2022language, jumper2021highly}, leverages ESM-2 to produce precise end-to-end structural predictions directly from a protein sequence. Figure~\ref{plms} provides further information on ESM-2, ESMFold, and AlphaFold.
ESMFold has demonstrated superior atomic-level predictions compared to AlphaFold2 when provided with a single sequence and has also delivered competitive performance when given complete Multiple Sequence Alignments (MSAs) as input.

Despite the impressive achievements in structure prediction from the sequence, pre-trained ProteinLMs remain exceptionally large, exceeding billions of parameters. Consequently, efficient deployment of these models on resource-constrained systems has become crucial due to their prohibitive memory footprint and energy consumption.

One practical approach to address this issue is neural network quantization~\cite{quantization_whitepaper,bai2022towards}. Quantization employs low-bit fixed-point arithmetic for weight and activation tensors instead of floating-point arithmetic, thereby reducing inference time and memory consumption~\cite{kim2021bert,bondarenko-etal-2021-understanding}.
Post-training quantization (PTQ) is more desirable than quantization-aware training (QAT) because it does not require retraining or access to the complete training dataset, which saves time and protects data privacy. However, quantization can introduce additional noise into the network, which may lead to a decrease in performance.

This paper investigates the PTQ technique for ESMFold, which is built upon the state-of-the-art ESM2 ProteinLM. Although neural network quantization has been extensively studied in CV and NLP domains~\cite{kim2021bert, PTQ4ViT_arixv2022, bondarenko-etal-2021-understanding,bai2022towards}, there has been no research on the quantization of ProteinLMs.
Our findings show that the widely used uniform quantization approach leads to a significant TM-Score~\cite{zhang2005tm} drop when applied to ESMFold. The main cause of this degradation is the wide and asymmetrical distribution of activation values before specific layers. Uniform quantization assigns equal quantization levels to the negative and positive activation components, resulting in significant prediction degradation when operating at low-bit widths.

To mitigate the aforementioned issues, we propose a new PTQ method for ProteinLMs (PTQ4Protein). Our contributions are as follows. 
(i) We demonstrate that the weight of ESMFold can be quantized to 8 bits with uniform quantization techniques, reducing memory usage by nearly 4$\times$ with negligible precision loss. However, quantizing the activations leads to significant degradation in prediction accuracy.
(ii) We conduct a systematic study to identify the underlying reason that prevents efficient activation quantization. The primary bottleneck is the wide and asymmetrical distribution of activation values before Layer Normalization. 
(iii) Based on our findings, we propose PTQ4Protein that uses piecewise linear quantization for asymmetrical activation values and achieves minimal accuracy loss. We verify PTQ4Protein's performance on the protein structure prediction task using the quantized ESMFold.
(iv) Furthermore, since ESMFold shares modules with other ProteinLM-based models, we supplement the experiment on the contact prediction task. The experimental results show that PTQ4Protein has the potential to significantly reduce the memory footprint and energy consumption of various ProteinLM-based applications.

\section{Related Work}

\subsection{Protein Language Models}
Large-scale language models~\cite{vaswani2017attention} with the self-supervised learning (SSL) paradigm, such as the masked language model (MLM)~\cite{kenton2019bert} and auto-regression~\cite{radford2018improving}, have achieved extraordinary success in NLP tasks. Recent progress has revealed that their capabilities are deeply related to the scale of the model parameters: the larger the scale of the parameters, the better the performance~\cite{brown2020language}.

Inspired by these achievements, large ProteinLMs using the Transformer~\cite{vaswani2017attention} architecture have been widely adopted~\cite{rives2021biological,rao2019evaluating}. 
An amino acid sequence, similar to the sequence of words or tokens in NLP, can represent a protein. Recent works have shown that by pre-training with only single sequences without much supervision, ProteinLMs such as ESM-1b~\cite{rives2019biological}, ESM-2~\cite{lin2022language} and HelixFold-Single~\cite{fang2022helixfold_single} have been successfully applied for various protein-related tasks, including secondary structure prediction~\cite{rao2019evaluating,rives2021biological}, contact prediction~\cite{rao2020transformer,rives2021biological}, 3D structure prediction~\cite{jumper2021highly,baek2021accurate}, annotation prediction~\cite{bileschi2019annotate}, protein-protein interaction prediction~\cite{evans2021protein}, and fitness prediction~\cite{dallago2021flip,hie2022evolocity}.

In this paper, we choose the representative protein structure prediction model, ESMFold, as the target for post-training quantization. ESMFold is based on the ESM-2 language model and is a protein structure prediction model to understand and encode protein sequences.

\begin{table*}
    \centering
    \caption{Quantization results and activation analysis on CASP14~\cite{kryshtafovych2021critical} and CAMEO~\cite{haas2018continuous} benchmarks. The metrics TM-Score can be found in~\cite{zhang2005tm}; in all cases, higher is better. The result in the bracket is the metric drop from floating-point networks. The top section of the table displays post-training quantization results, and the bottom section presents a leave-one-out analysis of activation quantization. The leave-one-out analysis was conducted using FP32 weights.}
    \begin{tabular}{llcc}
        \toprule
         Mode & Configuration & CASP14$\uparrow$ & CAMEO$\uparrow$ \\
         \midrule  
         \multirow{4}{*}{Full Quantization} & FP32 & 52.52 & 80.09 \\
         & W8A8 & 49.87 (-2.65) & 73.43 (-6.66) \\
         & W32A8 & 50.06 (-2.46) & 74.17 (-5.92) \\
         & W8A32 & 52.35 (-0.17) & 80.03 (-0.06) \\
         \midrule  
         \multirow{6}{*}{Activation Quantization} & only softmax input & 52.28 (-0.24) & 80.01 (-0.08) \\
         & only softmax output & 52.46 (-0.06) & 78.09 (-2.00) \\
         & only FFN (Linear) input & 51.77 (-0.75) & 79.80 (-0.29) \\
         & only LayerNorm output & 51.20 (-1.32) & 79.48 (-0.61) \\
         & only LayerNorm input & \textbf{50.34 (-2.18)} & \textbf{75.57 (-4.52)} \\
         & only LayerNorm input (layers 26-35) & \textbf{50.89 (-1.63)} & \textbf{76.24 (-3.85)} \\
         \bottomrule
    \end{tabular}
    \label{uniform_ablation_study}
\end{table*}

\subsection{Neural Network Quantization}
The efficient deployment of DNNs has been extensively studied in both academic and industrial settings. One of the most effective ways to reduce the computational time and memory consumption of neural networks is through quantization, which utilizes low-bit representations for weights and activation tensors. By storing model parameters in fewer bits and executing computation on integer arithmetic units instead of power-hungry floating-point ones, quantization significantly reduces memory overhead and computational cost. Specifically, when transitioning from 32 to 8 bits, the memory overhead of storing tensors decreases by a factor of 4, while the computational cost of matrix multiplication reduces quadratically by a factor of 16. 

There are two types of quantization methods, QAT~\cite{PACT_arxiv2018,quantization_whitepaper} and PTQ~\cite{PTQ_4bit_rapid_deployment_nips2019,bai2022towards}.
QAT methods combine quantization with network training. It optimizes quantization parameters to minimize task loss on a labeled training dataset.
PTQ methods quantize networks with a few unlabeled datasets, which is significantly faster than QAT and does not require any labeled datasets. 
PTQ methods should determine the scaling factors $\Delta$ and clipping range $[r_l, r_u]$ of activations and weights for each layer. One way of finding good quantization ranges is~\emph{static range estimation}, which determines quantization parameters for the network by passing a few batches of calibration data through the model before inference.
It yields more efficient inference because all quantization parameters are known in advance and fixed.

In this paper, we focus on PTQ methods that directly convert pre-trained full-precision models to their low-precision counterparts.

\section{Quantization Schemes}
In this section, we discuss three main parts. First, we introduce the standard uniform PTQ method. Second, we look into its issues and provide potential solutions. Lastly, we introduce our solution, PTQ4Protein, crafted to quantize ProteinLM weights and activations into low-bit widths.

\begin{figure*}[tbp]
    \centering
    \includegraphics[width = 0.9\textwidth]{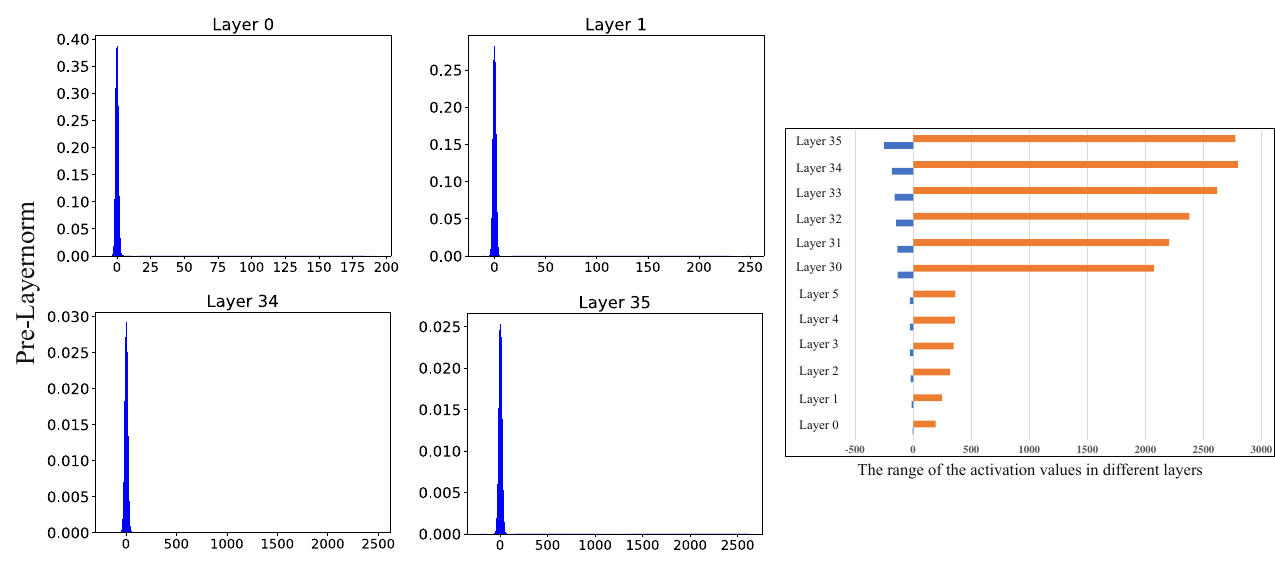}
    \caption{Distribution and Range of Pre-LayerNorm activation values in ESMFold across different layers. For the left part, the x-axis represents activation values, while the y-axis represents the proportion of values (as in Figure~\ref{post_layernorm}).}
    \label{pre_layernorm}
\end{figure*} 
\begin{figure*}
    \centering
    \includegraphics[width = 0.9\textwidth]{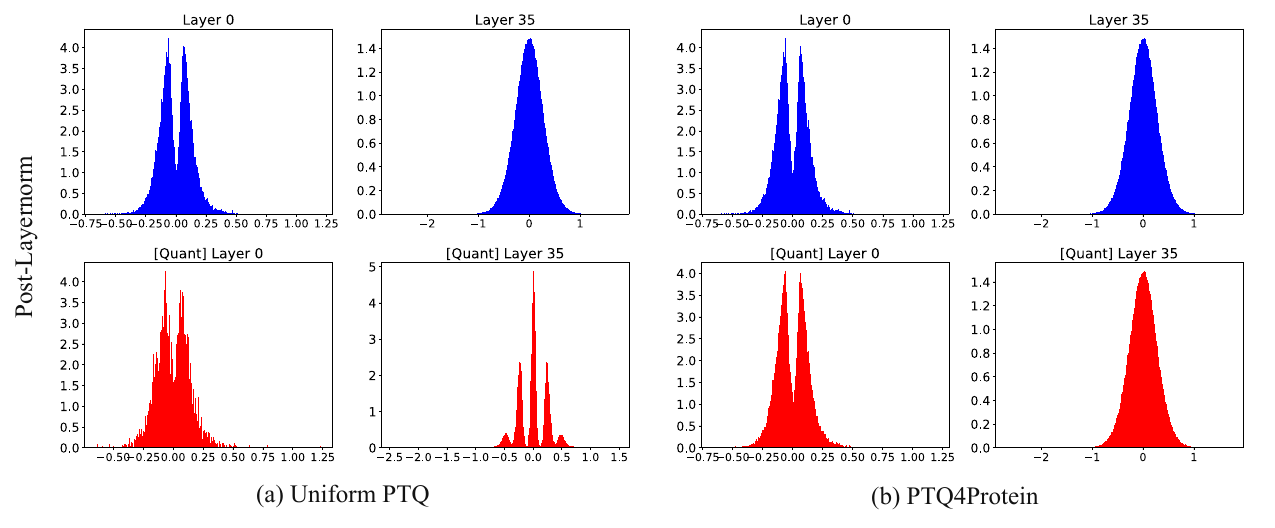}
    \caption{Distributions of Post-Layernorm activation values before and after quantization in ESMFold with uniform PTQ and PTQ4Protein.}
    \label{post_layernorm}
\end{figure*}
\subsection{Base Uniform Quantization}
Uniform quantization is a commonly used technique for quantization due to its practicality in implementing fixed-point arithmetic efficiently~\cite{krishnamoorthi2018quantizing}. It involves mapping real numbers $r$ with full precision to integer representations $r_q$ with lower precision using a linear transformation.

Based on \cite{choukroun2019low}, the approximated version $\hat{r}$ from uniform quantization scheme at $b$-bit is defined as: 
\begin{equation} 
\label{eq:uniform1}
\hat{r} =  \text{uni} (r; b, r_{l}, r_{u}, z)  =  s \times r_q + z
\end{equation}
\begin{equation}
\label{eq:uniform2}
r_q = \Bigl\lceil \frac{\text{clamp} (r; r_{l}, r_{u}) - z}{s} \Bigr\rfloor_{\mathbb{Z}_b}
\end{equation}
\begin{equation} 
\label{eq:uniform3}
\text{clamp}(r; r_{l}, r_{u}) = \min( \max(r, r_{l}), r_{u})
\end{equation}
\begin{equation} 
\label{eq:uniform4}
s = \frac{\Delta}{N-1}, \quad \Delta = r_{u} - r_{l}, \quad N=2^b
\end{equation}
where $[r_l, r_u]$ is the clipping range, $s$ is the scaling factor, $z$ is the offset, $N$ is the number of quantization levels, $r_q$ is the quantized integer computed by a rounding function $\lceil \cdot \rfloor_{\mathbb{Z}_b}$ followed by saturation to the integer domain $\mathbb{Z}_b$. We set the offset $z=0$ for symmetric signed distributions combined with $\mathbb{Z}_b = \lbrace -2^{b-1}, ..., 2^{b-1} -1 \rbrace$ and $z=r_l$ for asymmetric unsigned distributions (e.g., ReLU-based activations) with $\mathbb{Z}_b = \lbrace 0, ..., 2^{b} -1 \rbrace$.
To determine the range of activation values required for quantization, it is necessary to construct a calibration set from the original training data.

The optimization objective of most PTQ methods is to minimize the reconstruction error (REM), which refers to reducing the distance between the outputs of different operations carried out using quantized and full-precision counterparts~\cite{bai2022towards}. Therefore, the main focus of this paper is on REM-based PTQ methods.

\subsection{Limitations of Uniform Quantization}
The given definition states that uniform quantization divides the range evenly, regardless of the distribution of $r$. However, in practice, the weight and activation distributions of pre-trained DNNs often resemble bell-shaped Gaussian or Laplacian curves. Due to this, uniform quantization alone may not consistently minimize the approximation error, leading to a reduction in model accuracy.

In this study, we investigate the effects of standard 8-bit PTQ on ESMFold's performance. We employ uniform quantization with static range estimation to quantize fine-tuned models, quantizing all layer weights and activations. We report the best-performing configuration per benchmark, determined by TM-Score~\cite{zhang2005tm}, in the top section of Table~\ref{uniform_ablation_study}. Our results reveal a significant decrease in performance when applying joint 8-bit quantization, whereas weight quantization results in a minimal error. We observe that most of the degradation is due to activation quantization. 

To determine which part of the network is most affected by quantization, we conduct an ablation study where we selectively quantize activations in different modules. The results are presented in the bottom section of Table~\ref{uniform_ablation_study}. We observe the smallest performance drop when we avoid quantizing the activations preceding the Layer Normalization (LayerNorm). To better understand why quantizing the input to LayerNorm has a particularly detrimental effect on performance, we investigate the activation tensors in different layers. As shown in Figure~\ref{pre_layernorm}, we observe that each input to LayerNorm has a wide and asymmetric range. We hypothesize that applying uniform quantization for the LayerNorm input would result in significant errors due to the inherent trade-off between range and precision.
High dynamic ranges for small-range values lead to significant rounding errors, while a small dynamic range for large-range values results in high clipping errors. The significant difference in dynamic ranges presents a challenge for quantization. In Figure~\ref{post_layernorm}-(a), we present the distributions of Post-LayerNorm activation values before and after quantization. We observe that the quantization error in the LayerNorm input is amplified after the LayerNorm operation. Consequently, the dequantized activation values differ significantly from the original values, particularly in the final layers (e.g., Layer 34 and Layer 35).

\subsection{PTQ4Protein}
PTQ4Protein uses piecewise linear quantization to tackle activation quantization challenges in ESMFold, which is more complex due to the distribution of activation values. Previous research~\cite{pwlq} applied this method to resolve weight quantization problems in CV tasks. To split the quantization range, PTQ4Protein creates two regions, a dense central region, and a sparse high-magnitude region, and assigns an equal number of quantization levels ($N=2^b$) to each one. Our approach uses two regions with two breakpoints.

We divide the quantization range $[r_l, r_u]$ into two regions, the central region $R_1=[p_l,p_u]$ and the tail region $R_2=[r_l,p_l)\cup(p_u,r_u]$, using two breakpoints, $p_l$ and $p_u$.
After splitting the range, each region contains a negative and positive piece. We apply $(b-1)$-bit ($b \geq 2$) uniform quantization (as described in Section 3.1) within each of the three pieces so that every value, including the sign, is represented as $b$-bit within the quantization range. We define the $b$-bit piecewise linear quantization scheme as:
\begin{equation}
\label{eq:pw-scheme1}
\hat{r} = \text{pw}(r; b, r_l, r_u, p_l, p_u)= \left\{\begin{matrix}
\text{uni}(r_1; b-1, p_l, p_u, p_l) \\ 
\text{uni}(r_2; b-1, r_l, p_l, r_l) \\
\text{uni}(r_3; b-1, p_u, r_u, p_u)
\end{matrix}\right.
\end{equation}
where the full-precision real number $r$ is divided into three parts. The $r_1 \in [p_l, p_u]$, $r_2 \in [r_l, p_l)$, and $r_3 \in(p_u, r_u]$. 

\paragraph{Calibration Datasets}
To determine the suitable clipping range $[r_l, r_u]$, calibration datasets need to be sampled. In the case of protein structure prediction, amino acid sequences can vary significantly in length, ranging from dozens to hundreds of residues. Given the limited size of calibration datasets, meticulous preparation is vital to ensure that the range uniformly encompasses the distribution of sequence lengths. 

\paragraph{Breakpoint Selection}
To apply piecewise linear quantization, we require the optimal breakpoints $p$ to divide the quantization ranges into non-overlapping regions. Following work~\cite{pwlq}, we assume activations satisfy Gaussian distributions and use a simple and fast approximation as:
\begin{equation} 
\label{eq:breakpoint}
p_{l,u} = \ln(m\cdot r_{l,u} + n)
\end{equation}
The values of $p_l$ and $p_u$ are calculated based on $r_l$ and $r_u$, where $m$ and $n$ are hyperparameters used in the calculation. By default, the values of $m$ and $n$ are set as 0.8614 and 0.6079. 

\paragraph{Summary}
PTQ4Protein combines the benefits of both base uniform and piecewise linear quantization approaches. It can quantize both weights and activation values. PTQ4Protein uses the base uniform symmetric quantization method to quantify all layer weights, which has been shown to have a minimal error in previous studies. For activation quantization, PTQ4Protein applies the piecewise linear quantization method to maintain high precision while reducing memory and computation requirements.

To evaluate the effectiveness of PTQ4Protein, we visualized the distributions of Post-Layernorm activation values in ESMFold using our method, as shown in Figure~\ref{post_layernorm}-(b). We observed that the dequantized activation values were highly similar to the original values, indicating that PTQ4Protein can achieve effective quantization of ESMFold to low-bit widths with minimal loss. Detailed experimental results will be presented in the subsequent section.

\begin{table*}[t]
    \centering
    \caption{Quantization results for ESMFold on two protein tasks: supervised structure prediction and unsupervised contact prediction. We compare against the uniform quantization.}
    \begin{tabular}{ l|cc|c }
        \toprule
         Task & \multicolumn{2}{c|}{Supervised Structure Prediction} & Unsupervised Contact Prediction \\
         \midrule  
         Test Set & CASP14$\uparrow$ & CAMEO$\uparrow$ & ESM Data$\uparrow$ \\
         \midrule  
         FP32 baseline & 52.52 & 80.09 & 50.15 \\
         PTQ baseline, W8A8 & 49.87 (-2.65) & 73.43 (-6.66) & 46.27 (-3.88) \\
         PTQ baseline, W6A6 & 47.24 (-5.28) & 71.50 (-8.59) & 45.24 (-4.91) \\
         \midrule
         PTQ4Protein, W8A32 & 52.35 (-0.17) & 80.03 (-0.06) & 50.13 (-0.02) \\
         PTQ4Protein, W16A16 & 52.21 (-0.31) & 80.02 (-0.07) & 50.12 (-0.04) \\
         PTQ4Protein, W8A16 & 52.18 (-0.34) & 80.01 (-0.08) & 50.12 (-0.04) \\
         PTQ4Protein, W8A8 & \textbf{52.16 (-0.36)} & \textbf{80.01 (-0.08)} & \textbf{49.86 (-0.29)} \\
         PTQ4Protein, W6A6 & \textbf{52.03 (-0.49)} & \textbf{79.97 (-0.12)} & \textbf{49.24 (-0.91)} \\
         \bottomrule
    \end{tabular}
    \label{exp_v1}
\end{table*}
\begin{table*}[t]
    \caption{Ablation study of different modules in ESMFold with PTQ4Protein. Here the \textbf{FT} denotes Folding Trunk and \textbf{SM} denotes Structure Module.}
    \label{exp_v3}
    \centering
    \begin{tabular}{ lcccc }
        \toprule
         Quantization Parts & CASP14$\uparrow$ & CAMEO$\uparrow$ & GPU Memory Usage (GB) & Model Compress Ratio (\%) \\
         \midrule  
         FP32 baseline & 52.52 & 80.09 & 13.82 & 100\% \\
         \midrule
         Full, W8A8 & 52.16 (-0.36) & 80.01 (-0.08) & 3.45 & 25.0\%  \\
         Full, W6A6 & 52.03 (-0.49) & 79.97 (-0.12) & 2.58 & 18.8\%  \\
         \midrule
         Only ESM-2, W8A8 & 52.46 (-0.06) & 80.06 (-0.03) & 5.47 & 39.6\%  \\
         Only ESM-2, W6A6 & 52.42 (-0.10) & 79.97 (-0.12) & 4.76 & 34.5\%  \\
         Only FT and SM, W8A8 & 52.20 (-0.32) & 80.02 (-0.07) & 11.78 & 85.3\%  \\
         Only FT and SM, W6A6 & 52.11 (-0.41) & 79.58 (-0.51) & 11.62 & 84.1\%  \\
         \bottomrule
    \end{tabular}
\end{table*}

\section{Experiments}
In this section, we first present experiment settings. Secondly, we evaluate PTQ4Protein by conducting experiments on two protein tasks. Finally, we perform an ablation study to analyze the impact of the quantization method on different components of ESMFold.

\subsection{Experiment Settings}
\paragraph{Model and Dataset}
We chose ESMFold, which is representative of ProteinLM-based models, as our quantization target. The version of ESMFold we used was released in 2022-11 with 3B parameters\footnote{https://github.com/facebookresearch/esm}.
To evaluate the effectiveness of our quantization method, we utilized datasets from two prominent competitions, namely CASP14~\cite{kryshtafovych2021critical} and CAMEO~\cite{haas2018continuous}. Specifically, for CASP14, we extracted domain-level targets from the Free-Modeling and Template-Based Modeling hard categories, considering only contiguous domains that form part of protein chains deposited in the PDB. This yielded a total of 34 target domains, with a maximum sequence length of 405 residues. On the other hand, for CAMEO, we collected a set of 143 targets released between March and May 2022, encompassing samples from all three difficulty levels (easy, medium, and hard) and with a maximum length of 700 residues.

To construct our calibration datasets, we selected an additional 100 targets from the CASP13 competitions. These targets were chosen such that their length range is uniformly distributed across the original length distributions.
As an additional experiment, we evaluate our quantization method on OmegaFold~\cite{OmegaFold}, another well-known ProteinLM-based model that shares a similar structure with ESMFold. Moreover, we conduct quantization experiments on the unsupervised contact prediction task, employing ESM-2~\cite{lin2022language} to learn contacts within self-attention maps. We used evaluation data provided by Facebook.

\paragraph{Evaluation Metrics}
For the supervised structure prediction task, we consider the TM-score~\cite{zhang2005tm}, widely used in previous works and CASP challenges. For the unsupervised contact prediction task, we evaluate the precision of the top $L$ contacts~\cite{rao2020transformer}.

\begin{table}
    \caption{Quantization results for OmegaFold on protein structure prediction task}
    \begin{center}
    \begin{tabular}{lccc}
        \toprule
         Model & Method & CASP14$\uparrow$ & CAMEO$\uparrow$ \\
         \midrule
         OmegaFold,  FP32 & -- & 53.04 & 76.11 \\
         OmegaFold,  W8A8 & PTQ baseline & 48.88 (-4.16) & 70.23 (-5.88) \\
         OmegaFold,  W8A8 & PTQ4Protein & \textbf{52.90 (-0.14)} & \textbf{75.95 (-0.16)} \\
         \bottomrule
    \end{tabular}
    \end{center}
    \label{exp_v4}
\end{table}
\begin{table}
    \caption{Quantization results on ESMFold with different sampling modes for calibration datasets.}
    \begin{center}
    \begin{tabular}{lcc}
        \toprule
         Sampling Mode & CASP14$\uparrow$ & CAMEO$\uparrow$ \\
         \midrule
         Random,  W8A8 & 51.95 & 79.01 \\
         Prefer Short,  W8A8 & 50.14 & 78.86 \\
         Prefer Long,  W8A8 & 52.03 & 79.32 \\
         Uniform,  W8A8 & \textbf{52.16} & \textbf{80.01} \\
         \bottomrule
    \end{tabular}
    \end{center}
    \label{exp_v2}
\end{table}
\subsection{Experiment Results}
We compared different quantization settings for ESMFold using PTQ4Protein and summarized the results in Table \ref{exp_v1}. Our findings indicate that at 8-bit and 6-bit quantization levels, PTQ4Protein outperforms the base uniform quantization method, resulting in improved accuracy in protein structure and contact predictions.

The results in Table \ref{exp_v1} (left part) indicate that the base PTQ method exhibits a significant drop in TM-Score of nearly 3\% and 7\% on the CASP14 and CAMEO benchmarks when using 8-bit quantization. In contrast, our PTQ4Protein method achieves a much smaller loss of less than 0.4\% and 0.1\% on the CASP14 and CAMEO benchmarks, respectively. 
Similarly, when using 6-bit quantization, the base PTQ method results in a considerable accuracy drop of 5.28\% on CASP14 and 8.59\% on CAMEO, whereas our PTQ4Protein method achieves a much smaller accuracy drop of 0.49\% on CASP14 and 0.12\% on CAMEO. These findings demonstrate the effectiveness of our proposed PTQ4Protein method and suggest that ESMFold can be quantized to relatively low-bit widths (e.g., 6-bit) with minimal loss in accuracy. 
Since ESMFold shares modules with other ProteinLM-based models, we supplement quantization experiments on ESM-2 and OmegaFold.
The results on the contact prediction task (right part of Table \ref{exp_v1}) and OmegaFold (Table \ref{exp_v4}) further verify the effectiveness of PTQ4Protein.

\subsection{Ablation Study}
We conducted an ablation study to investigate the impact of different modules in ESMFold.
Based on statistical results, the ESM-2 component within ESMFold constitutes the majority of its parameters (80.4\%), while the Folding Trunk (FT) and Structure Module (SM) comprise only a small fraction of the parameters (19.5\%). 
Table \ref{exp_v3} presents the experimental results obtained.
As we can see, the Folding Trunk and Structure Module are more sensitive to quantization than the ESM-2.
With PTQ4Protein, quantizing the Structure Module results in a higher TM-Score drop (0.26\% in 8-bit and 0.31\% in 6-bit on CASP14, and correspondingly 0.04\% and 0.39\% on CAMEO) than quantizing the ESM-2. This experimental result shows that the part of ESM-2 is more robust and more accessible to be quantized, even in very low-bit widths.

Table \ref{exp_v3} also shows the comparison of GPU memory usage and model compress ratio. 
PTQ4Protein effectively conserves the GPU memory occupied by the model. 
This is particularly interesting given that the ESM-2 is a general-purpose ProteinLM that has demonstrated utility in various biological unsupervised and supervised tasks~\cite{lin2022language}. Our findings suggest that PTQ4Protein might have a broader scope beyond the realm of protein structure prediction. It holds the potential for application in various other protein-related tasks that involve predicting protein functions or other inherent properties derived from individual sequences.
\begin{figure*}[t]
  \centering
  \begin{subfigure}[b]{0.2\textwidth}
  \includegraphics[width=\textwidth]{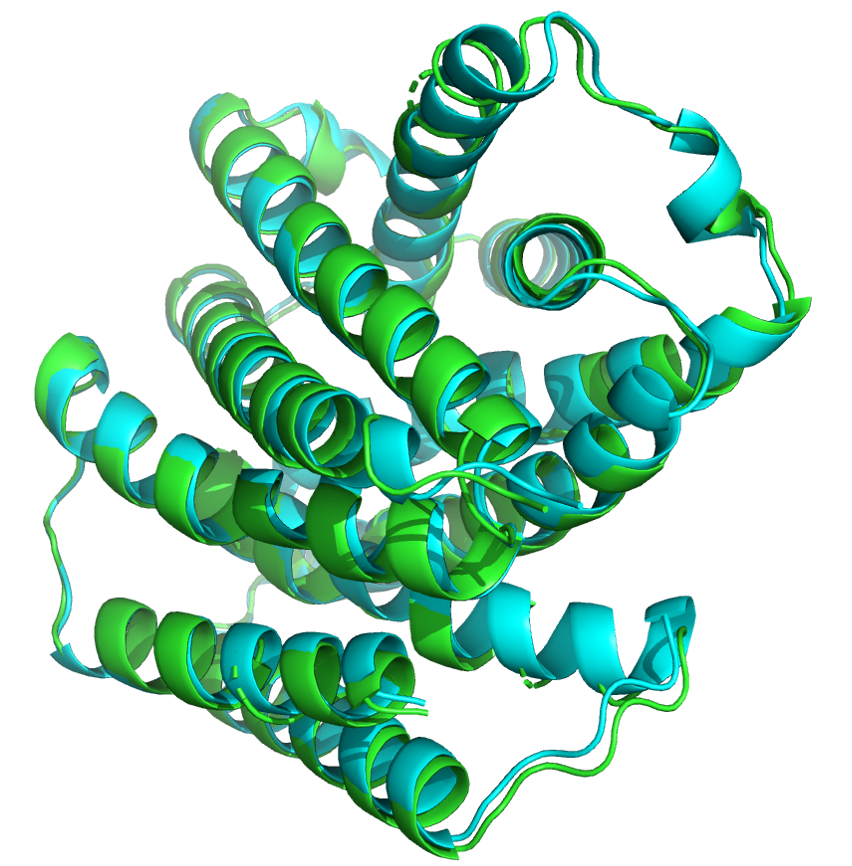}
  \caption{7B0K,Uniform\\TM-score=0.7980}
  \end{subfigure}
  \hspace{0.1cm}
  \begin{subfigure}[b]{0.2\textwidth}
  \includegraphics[width=\textwidth]{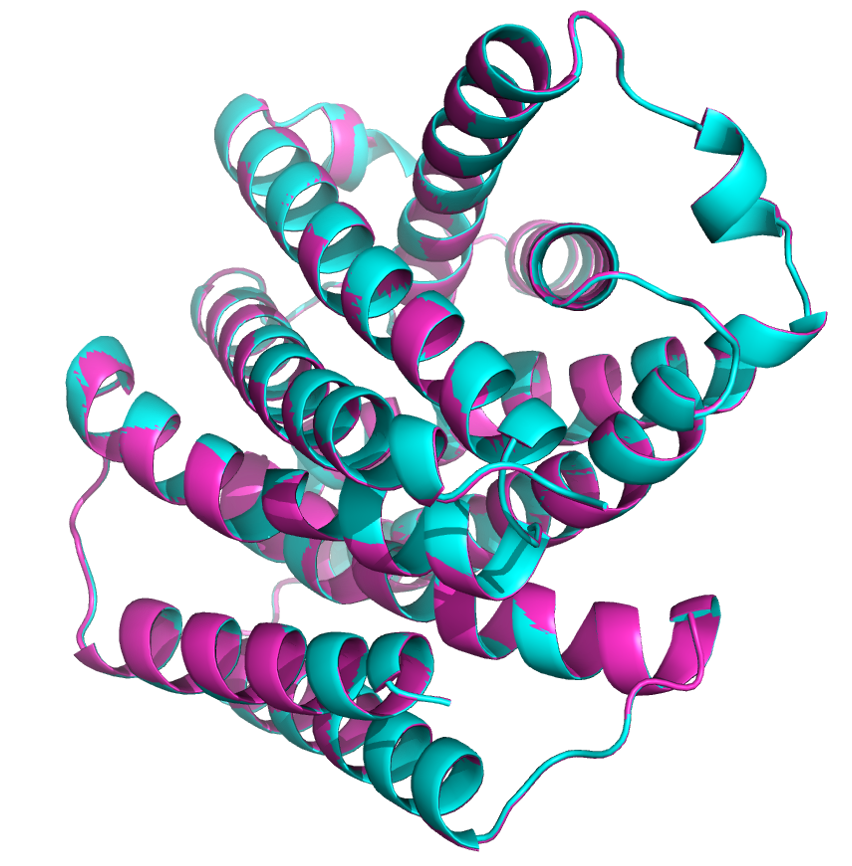}
  \caption{7B0K,PTQ4Protein\\TM-score=0.8095}
  \end{subfigure}
  \hspace{0.3cm}
  \begin{subfigure}[b]{0.2\textwidth}
  \includegraphics[width=\textwidth]{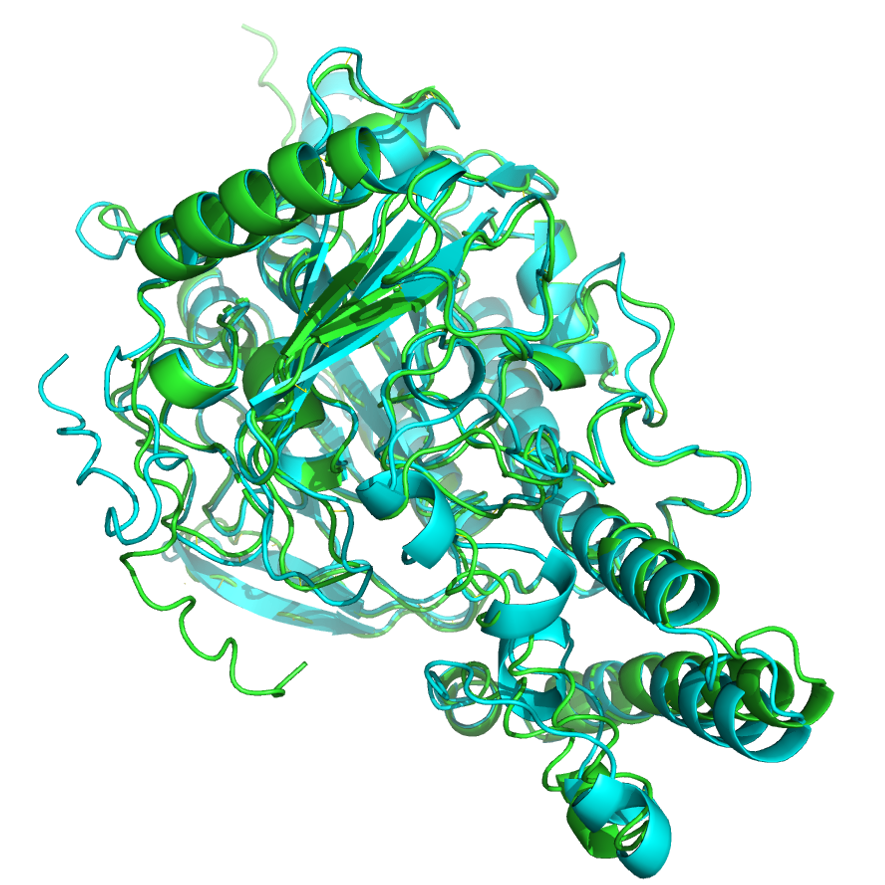}
  \caption{7EBQ,Uniform\\TM-score=0.9178}
  \end{subfigure}
  \hspace{0.1cm}
  \begin{subfigure}[b]{0.2\textwidth}
  \includegraphics[width=\textwidth]{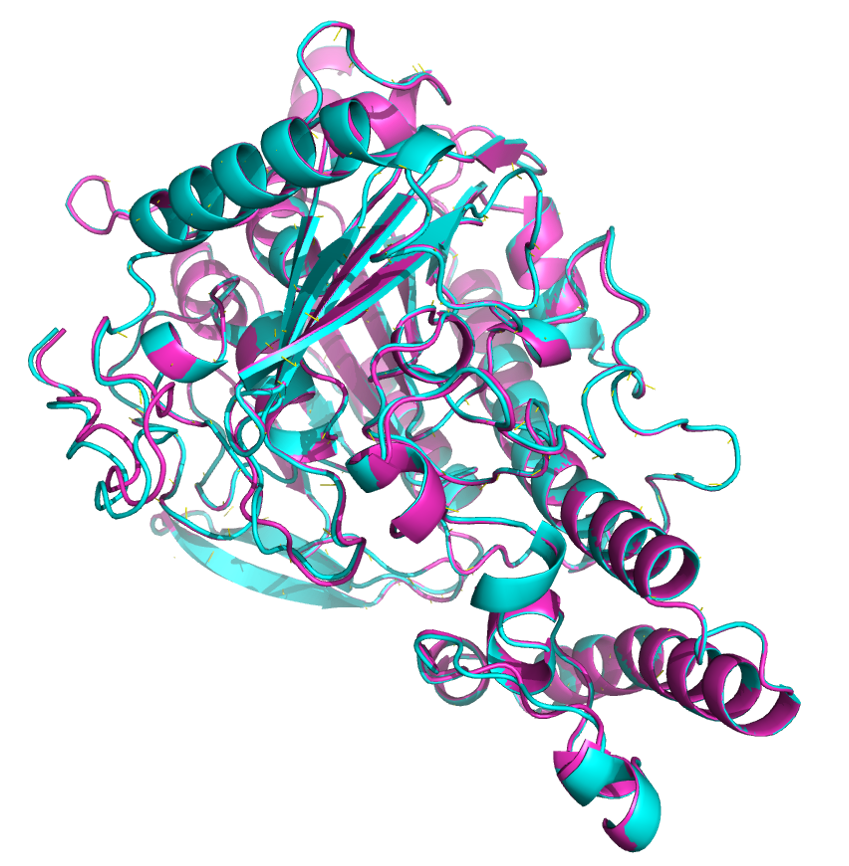}
  \caption{7EBQ,PTQ4Protein\\TM-score=0.9659}
  \end{subfigure}
  \caption{Alignment of Predicted 3D Structures of Proteins Using Uniform PTQ and PTQ4Protein with Reference Structure (PDB ID: 7B0K and 7EBQ). Green indicates the structure predicted by Uniform PTQ, magenta indicates the structure predicted by PTQ4Protein, and cyan represents the reference structure.}
  \label{structure}
\end{figure*}
\subsection{Calibration Strategies and Case Study}
We compared the following sampling modes for constructing the calibration datasets. (1) \emph{Random}: randomly selecting calibration data, (2) \emph{Prefer Short}: preferring shorter sequence data, and (3) \emph{Prefer Long}: preferring longer sequence data. Notably, all sampling modes have an equal calibration data size.
The experimental results are presented in Table \ref{exp_v2}.

The experimental results show that the performance of quantized ESMFold is affected by the choice of sampling modes for constructing the calibration datasets. Specifically, the \emph{Random} and \emph{Prefer Short} sampling modes result in poorer performance, indicating that an appropriate clipping range cannot be obtained from shorter calibration data. On the other hand, our proposed uniform sampling mode performs better than the \emph{Prefer Long} mode. This finding suggests that a uniform sampling mode, which covers the complete length distributions, is more suitable for calibrating ESMFold.

In addition, we examined the effectiveness of PTQ4Protein on two proteins from the CASP14 dataset by directly visualizing the predicted 3D structures. Figure \ref{structure} displays the predicted 3D structures for Uniform PTQ and PTQ4Protein. Our results indicate that PTQ4Protein predicted structures that were closer to the native structures.


\section{Conclusion and Future Work}
In this study, we investigated the application of quantization to the representative ProteinLMs and identified unique challenges associated with their highly asymmetric activation ranges. Specifically, representing these ranges using low-bit fixed-point formats was found to be difficult. To solve this problem, we propose PTQ4Protein, a novel method for accurate PTQ of ProteinLMs.
PTQ4Protein enables efficient quantization in low-bit widths with minimal loss. These results indicate its potential for facilitating the rapid deployment of biological science applications on resource-limited devices.

In the future, we intend to apply our proposed method to other protein-related tasks, including fitness and protein-protein interaction predictions. Additionally, we plan to extend our approach to other variations of ProteinLMs and further explore compression techniques for these networks.

\section{Acknowledgement}
This work is supported by National Natural Science Foundation of China Grant (No. U22A6001) and Key Research Project of Zhejiang Lab (No. 2022PG0AC02, No. K2023NB0AC11).

\bibliographystyle{abbrv}
\bibliography{Reference}

\end{document}